\documentclass[final,3p,times,twocolumn]{elsarticle}

\usepackage[linesnumbered,ruled,vlined]{algorithm2e}
\usepackage{blindtext, graphicx, amsmath, algpseudocode, pifont, algcompatible, comment, layout, amsthm, amssymb, subfigure, subcaption}
\usepackage{enumitem}   
\usepackage{eso-pic}
\usepackage{booktabs}
\usepackage{float}
\usepackage{multirow,multicol}
\usepackage{adjustbox}
\usepackage{tcolorbox}
\usepackage{color, colortbl}
\usepackage{soul}

\newcommand{\framework}{{\textbf{\textit{FedUNet}}}}

\newcommand{\rev}[1]{{\textcolor{black}{#1}}}

\usepackage[utf8]{inputenc}
\usepackage[english]{babel}
\usepackage{hyperref} 
\hypersetup{ colorlinks=true, linkcolor=black, filecolor=black, urlcolor=cyan, }

\usepackage{caption}
\journal{ICT Express}

\begin{document}

\begin{frontmatter}

\title{FedUNet: A Lightweight Additive U-Net Module for Federated Learning with Heterogeneous Models}

\author[1]{Beomseok Seo}
\author[2]{Kichang Lee}
\author[1]{JaeYeon Park\corref{cor1}}


\address[1]{Department of Mobile Systems Engineering, PRIMUS International College, Dankook University, Yong-in, Korea}
\address[2]{School of Integrated Technology, College of Computing, Yonsei University, Seoul, Korea}

\cortext[cor1]{Corresponding author. Email: jaeyeon.park@dankook.ac.kr}

\begin{abstract}

Federated learning (FL) enables decentralized model training without sharing local data. However, most existing methods assume identical model architectures across clients, limiting their applicability in heterogeneous real-world environments. To address this, we propose \framework{}, a lightweight and architecture-agnostic FL framework that attaches a U-Net-inspired additive module to each client’s backbone.
By sharing only the compact bottleneck of the U-Net, \framework{} enables efficient knowledge transfer without structural alignment. The encoder-decoder design and skip connections in the U-Net help capture both low-level and high-level features, facilitating the extraction of client-invariant representations. This enables cooperative learning between the backbone and the additive module with minimal communication cost. Experiment with VGG variants shows that \framework{} achieves 93.11\% accuracy and 92.68\% in compact form (i.e., a lightweight version of \framework{}) with only 0.89 MB low communication overhead.

\end{abstract}




\end{frontmatter}

\section{Introduction}
\label{sec:intro}
\vspace{-2ex}

Federated Learning (FL) enables privacy-preserving machine learning by training models collaboratively across decentralized clients without direct access to local data. This paradigm is valuable in domains like mobile computing and healthcare, where data is privacy-sensitive and distributed across heterogeneous devices.
%
%
%
Notably, conventional FL algorithms, such as FedAvg~\cite{mcmahan2017communication}, assume identical model architectures across all clients. While this simplifies parameter averaging, it limits practicality in real-world settings where clients differ in computational resources and task requirements. As a result, uniform-architecture FL methods struggle to accommodate model heterogeneity.

In recent years, heterogeneous federated learning, which permits each client to use a model architecture tailored to its computational resources or application requirements, has attracted considerable attention. However, this flexibility poses three main challenges: (a) model parameters cannot be directly aggregated due to the structural mismatches among client models~\cite{mcmahan2017communication, li2020federated}; (b) Most of existing approaches rely on costly knowledge-distillation techniques requiring additional shared datasets~\cite{li2019fedmd}, synthetic data generation via generative models, logit matching on large public corpora, feature-map alignment across heterogeneous networks, or ensemble-based teacher aggregation~\cite{liang2020think} which erode the efficiency gains of federated learning; (c) Achieving interoperability among structurally diverse models while preserving privacy and minimizing communication overhead remains challenging~\cite{baltruvsaitis2018multimodal,tan2022fedproto}. Specifically, FL methods for heterogeneous models often require the exchange of intermediate representations such as logits or feature maps, or involve multi-round distillation procedures based on shared or synthetic data. These approaches introduce significant communication costs, which undermine the scalability of federated learning systems and hinder their practical deployment in bandwidth-constrained or resource-limited environments. 


To address the challenges of federated learning with heterogeneous model architectures, we propose a novel framework, \framework{}, that incorporates a U-Net-inspired additive module into each client’s model. Unlike existing approaches that require structural uniformity or expensive distillation procedures, our method allows each client to retain its own backbone (e.g., VGG, ResNet) while attaching an auxiliary U-Net module. Importantly, only the bottleneck of the U-Net is shared and synchronized across clients, serving as a compact and architecture-agnostic medium for knowledge transfer. This enables efficient federated updates without aligning the main model architectures.

Specifically, the shared bottleneck plays a central role in harmonizing the diverse representations produced by different backbones. It serves as a lightweight representation hub aligning heterogeneous features into a unified latent space, facilitating cross-client interoperability. By capturing client-invariant and generalizable patterns, the bottleneck enables knowledge sharing even among clients with vastly different data distributions or model designs. Furthermore, skip connections in U-Net allow integrating low-level and high-level features, encoding fine-grained local information often overlooked by base models. This architectural synergy results in cooperative feature learning, where the additive module and backbone complement each other to enhance overall performance. At the same time, communication efficiency is preserved by limiting synchronization to the compact bottleneck, making the approach suitable for resource-constrained real-world deployment.

For evaluation, our experiments in a heterogeneous setting involving VGG and ResNet backbones show that our method can achieve comparable or improved accuracy with minimal computational overhead and significantly reduced communication cost. Specifically, despite the model diversity, our additive U-Net module achieves up to \textbf{93.11}\% accuracy by sharing only \textbf{56.66} MB. Notably, even a compact version of the U-Net enables efficient knowledge transfer with negligible increase in training time (\textbf{1.051} $\mathbf{\times}$ compared to FedAvg) and achieves \textbf{92.68}\% accuracy with only \textbf{0.89} MB of communication per round, offering a highly efficient trade-off. These results suggest that the proposed framework can serve as a practical and scalable solution for real-world federated systems characterized by model heterogeneity and device diversity. The main contributions of this paper are threefold.

\noindent{$\bullet$} We propose a federated learning framework, \framework{}, that allows clients to maintain their own model architectures by attaching a U-Net-based additive module. By sharing only the bottleneck, \framework{} enables architecture-independent collaboration without model alignment or a shared dataset.


\noindent{$\bullet$} The shared bottleneck acts as a lightweight hub aligning heterogeneous features into a common latent space. It captures client-invariant, generalizable knowledge, enabling effective cross-model knowledge transfer.


\noindent{$\bullet$} Utilizing the encoder-decoder structure and skip connections of U-Net, the additive module captures fine-grained, hierarchical features that complement the base model. This design supports cooperative learning with low communication and computational overhead.

\vspace{-2ex}
\section{Related Work and Motivation}
\label{sec:related}
\vspace{-2ex}

\subsection{Federated Learning for Heterogeneous Models}
While Federated Learning (FL) has been extensively studied in the context of homogeneous model settings, its extension to heterogeneous architectures has introduced new and significant challenges. Early FL approaches such as FedAvg~\cite{mcmahan2017communication} require all clients to share an identical model architecture, allowing simple parameter averaging but severely limiting applicability in environments with diverse device capabilities and model preferences.

To address heterogeneity of the model architecture, recent studies have explored model-agnostic federated learning and knowledge distillation-based FL frameworks~\cite{li2020federated,li2019fedmd}. These methods allow clients to train with different models, yet introduce their own set of limitations. A major issue is the lack of parameter compatibility, which prevents straightforward aggregation across models. To compensate, knowledge distillation techniques have been used to transfer knowledge via soft predictions or intermediate representations. However, these often require additional public datasets and entail high computational cost, such as using generative models for synthetic data generation, logit matching over large datasets, feature alignment across distinct architectures, or ensemble-based distillation~\cite{mohri2019agnostic,park2024fedhm}.

Furthermore, ensuring interoperability among structurally dissimilar models while maintaining privacy and communication efficiency remains an open problem. Methods that attempt to bridge model differences typically suffer from either architectural constraints or scalability issues in real-world deployment. These limitations highlight the urgent need for a communication-efficient, lightweight, and architecture-independent approach to enable effective federated learning across heterogeneous clients.

\vspace{-2ex}
\subsection{Design Principles of the Additive U-Net Module}
Our proposed framework, \framework{}, is motivated by the need for a lightweight and flexible mechanism that enables knowledge sharing across heterogeneous client models without requiring architectural alignment or costly knowledge distillation. While recent works have explored the use of U-Net architectures in federated learning settings~\cite{ronneberger2015u}, they have primarily focused on homogeneous model scenarios or domain-specific tasks such as segmentation. To the best of our knowledge, no prior work has utilized a U-Net-based module as a client-invariant feature extractor specifically designed to address architectural heterogeneity in federated learning. We define client-invariant features as representations that generalize across clients with different local data distributions, a characteristic essential for federated learning under heterogeneity.

To this end, we introduce an additive U-Net module that is attached to each client’s backbone model. Only the bottleneck of this module is shared and synchronized across clients, while the backbone remains entirely local. This bottleneck plays a crucial role as it transforms model-specific feature maps into a compact latent representation that facilitates interoperability among diverse architectures~\cite{koh2020concept}. By encoding high-dimensional features into a shared low-dimensional space, the bottleneck acts as a central hub for representation harmonization, enabling collaboration between structurally incompatible models.

Furthermore, the shared bottleneck is designed to capture client-invariant and transferable patterns that generalize across client datasets~\cite{koh2020concept}. While the backbone model focuses on learning data-specific representations, the U-Net bottleneck aggregates broadly useful information, such as edge or texture patterns, that benefit all clients. This shared knowledge enhances generalization without requiring a shared dataset or soft-label transfer. Especially, the encoder-decoder structure and skip connections within the U-Net module also allow it to preserve both low-level spatial features and high-level semantics~\cite{ronneberger2015u}. As a result, the module complements the local backbone by extracting fine-grained information that may otherwise be underutilized. This creates a cooperative feature learning process between the base model and the additive module.

Importantly, by limiting synchronization to only the bottleneck parameters, our approach significantly reduces communication overhead~\cite{park2024fedhm}. The bottleneck serves as a minimal yet sufficient unit of transferable knowledge, allowing the system to avoid the heavy costs typically associated with full model exchange or complex distillation pipelines~\cite{ronneberger2015u}. In summary, the additive U-Net module provides an efficient and generalizable foundation for federated learning in heterogeneous settings, enabling practical deployment without compromising model diversity or performance.
\section{\framework{}}
\label{sec:frame}
\vspace{-2ex}

\begin{figure}[t]
    \centering
    \includegraphics[width=1\columnwidth]{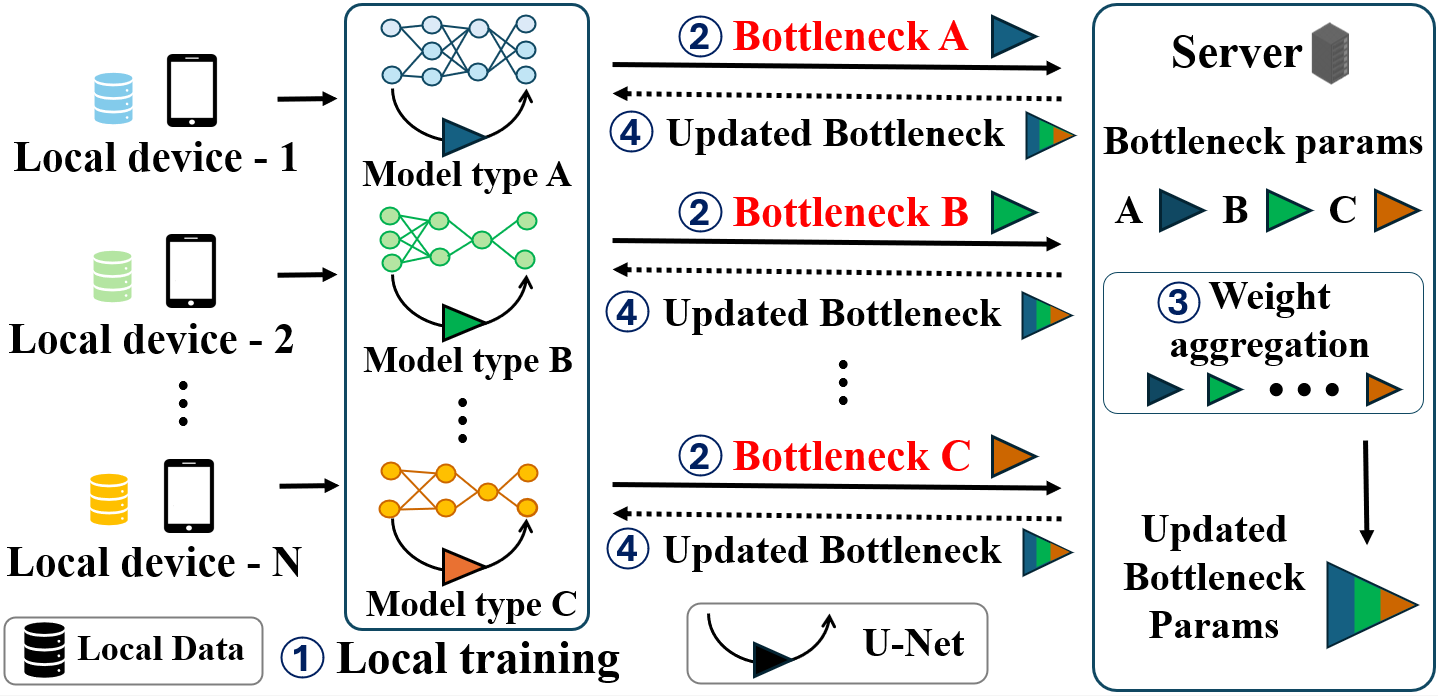}
    \vspace{-3ex}
    \caption{Overall architecture of \framework{}}
    \label{fig:overall_fedunet}
    \vspace{-3ex}
\end{figure}

To implement federated learning across heterogeneous model architectures, we propose \framework{}, a framework that introduces an additive U-Net module into each client model. As shown in Figure~\ref{fig:overall_fedunet}, \framework{} enables clients with different backbone architectures (e.g., VGG, ResNet) to collaborate by sharing only the compact bottleneck of their U-Net modules. The U-Net is connected in parallel to the client model and processes internal representations extracted from early and late layers. The bottleneck serves as a shared client-invariant encoder, while the rest of the model remains private and task-specific. This section details the design and functionality of the U-Net module and its integration with backbone models. We provide the overall workflow in Algorithm~\ref{alg:fedunet}.

\vspace{-2ex}
\subsection{Additive U-Net as a Client-Invariant Feature Extractor}

Specifically, the additive U-Net module receives the original input data $\mathbf{x}$ in parallel with the backbone model. While the backbone processes $\mathbf{x}$ through its standard layers to produce a task-specific feature map $\mathbf{f}_{\text{base}}$, the U-Net encodes $\mathbf{x}$ to produce a complementary representation $\mathbf{f}_{\text{unet}}$. This design allows both modules to extract features independently, enabling diverse information pathways without architectural interference.

The encoder performs spatial downsampling and deep feature extraction using a sequence of convolutional layers. The resulting representation is compressed into a low-dimensional bottleneck vector $\mathbf{z} \in \mathbb{R}^{d}$. Here, we define the `bottleneck' as the model component that computes $\mathbf{z}$, and the trainable parameter $w^{\mathbf{z}}$ is the only component synchronized across clients via parameter aggregation. This bottleneck acts as a shared representation space that aligns semantically rich (i.e., task-relevant) and architecture-independent features from all clients. It enables interoperability between heterogeneous models without requiring architectural uniformity or shared datasets.

Once the encoder produces $\mathbf{z}$, the decoder reconstructs spatially aligned features using upsampling layers, incorporating skip connections from the encoder to preserve low-level details. The output of the decoder is a feature map $\mathbf{f}_{\text{unet}} \in \mathbb{R}^{H \times W \times C}$, which is designed to be in the same shape as the backbone’s final feature map $\mathbf{f}_{\text{base}} \in \mathbb{R}^{H \times W \times C}$. These two outputs are then merged into a unified feature map $\mathbf{f}_{\text{joint}}$, defined as $\mathbf{f}_{\text{joint}} = \mathbf{f}_{\text{base}} + \mathbf{f}_{\text{unet}}$ or $\text{Concat}(\mathbf{f}_{\text{base}}, \mathbf{f}_{\text{unet}})$, depending on the fusion strategy. This joint feature is passed through the final fully connected layer to produce the prediction $\hat{y}$. Importantly, the entire model, including both the backbone and the additive U-Net module, is trained using a single classification loss. Letting $y$ denote the ground-truth label and $\hat{y}$ the final prediction, the training objective is $\mathcal{L} = \text{CrossEntropy}(\hat{y}, y)$.

During federated training, only the parameters of the bottleneck component $w^{\mathbf{z}}$ are uploaded and aggregated at the server side. This yields an updated global bottleneck, which is then downloaded by all clients for the next round. The rest of the model, including the client-specific backbone and the non-bottleneck layers of the U-Net, remains local.

This architecture allows \framework{} to decouple task-specific feature learning from globally transferable representation extraction. The backbone model remains fully customizable and optimized for local data, while the additive U-Net focuses on extracting and aligning general features that can be shared across clients efficiently. As a result, the system supports heterogeneous model architectures while maintaining low communication overhead and strong generalization capability.

\vspace{-2ex}
\subsection{Integration with Backbone Architectures}

\begin{figure}[t]
    \centering
    \includegraphics[width=1\columnwidth]{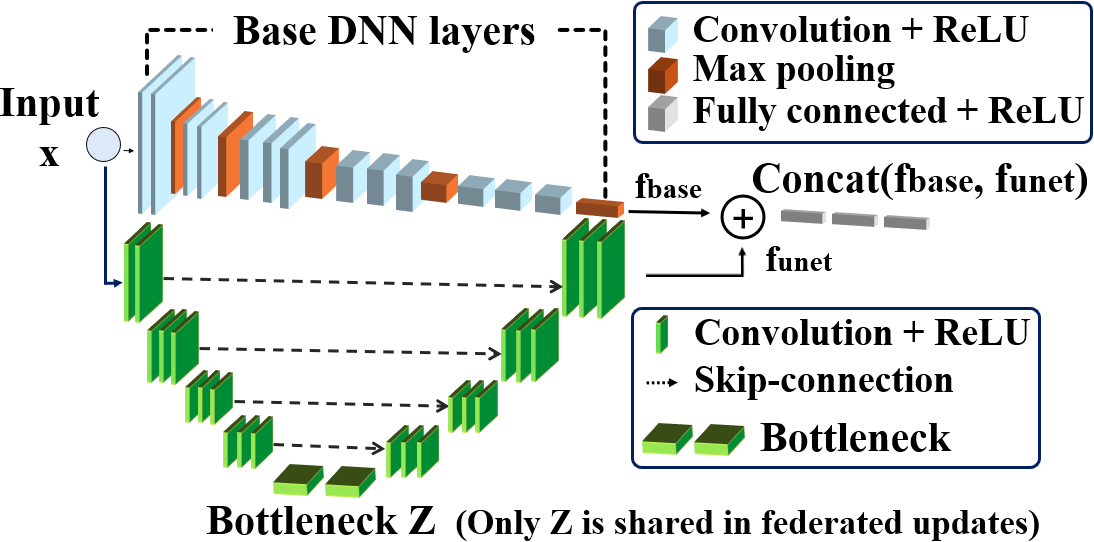}
    \vspace{-4ex}
    \caption{Sample connections of the additive U-Net module to a baseline DNN model.}
    \label{fig:fedunet_exam}
    \vspace{-3ex}
\end{figure}

\begin{algorithm}[t]
\footnotesize
\newcommand\mycommfont[1]{\footnotesize\ttfamily\textcolor{blue}{#1}}
\SetCommentSty{mycommfont}

\SetKw{KwData}{Data:}
\SetKw{KwServer}{Server executes:}
\SetKw{KwParallel}{in parallel}
\SetKw{KwClient}{ClientUpdate ($k$, $\omega^Z$):}

\caption{\framework{} operations}
\label{alg:fedunet}

\DontPrintSemicolon

  \KwData{} $(D_{1}, ..., D_{N})$ where $D_{k}$ is the $k$-th local dataset. \;
  
  \KwServer{}

  \For{each round $r = 1, 2, ..., R$} {
    
    $S_{n} \leftarrow$ Random set of $n$ clients\;

    \tcp{$\omega^{Z}$; an additive U-Net module's bottleneck}
    \For{each client $k \in S_{n}$ \KwParallel{}} {
        $\omega_k^{Z} \leftarrow$ ClientUpdate($k$, $\omega^{Z}$)\;
    }
    
    \tcp*{Aggregate bottleneck parameters across selected clients}
    $\omega^{Z} \leftarrow \frac{1}{n} \sum_{k \in S_{n}} \omega_k^{Z}$\;
    
    \Return $\omega^{Z}$ \tcp*{Transmit updated bottleneck parameters to clients in round $r$}
  }

  \KwClient{}
    
  Split $D_k$ into mini-batches $\mathcal{B}$ of size $B$ \textit{(batch size)}
  
  Initialize local model $\omega_k = [\text{baseline DNN}, \text{additive U-Net module}]$;

  \For{each local epoch $i = 1$ to $E$ \text{(local epochs)}}{
    \For{each batch $b \in \mathcal{B}$} {
      $\omega_k \leftarrow \omega_k - \eta \nabla \ell(\omega_k; b)$\;
    }
  }
  
  \Return $\omega_k^{Z}$ \tcp*{Return trained bottleneck parameters only}
\end{algorithm}

The additive U-Net module in \framework{} is designed to be integrated seamlessly into existing client models without requiring structural modifications. This design ensures compatibility with a wide range of architectures and preserves the original functionality and optimization flow of the backbone model. Figure~\ref{fig:fedunet_exam} illustrates how the U-Net module is attached to a typical VGG-style convolutional neural network. In this configuration, the U-Net module receives the original input data $\mathbf{x}$ in parallel with the backbone model, rather than intermediate features from it. The backbone proceeds through its standard layers to produce the final feature map $\mathbf{f}_{\text{base}}$, while the U-Net processes $\mathbf{x}$ to produce an auxiliary feature map $\mathbf{f}_{\text{unet}}$. These two representations are then concatenated or combined before entering the final fully connected layer. This design supports modular integration without altering the internal structure of the backbone network.

This structure enables the model to extract features through two parallel paths. The backbone network continues to learn task-specific representations optimized for local data, while the additive U-Net focuses on learning generalizable patterns that can be shared across clients. The architecture allows for cooperative learning between the base and the U-Net without interfering with each other’s gradients or parameter spaces.

Furthermore, to support flexible deployment across diverse devices, the fusion mechanism between the U-Net output and the main model output can be implemented in various ways. For example, simple concatenation, addition, or even learnable fusion layers can be employed, depending on the computation budget of the client. Since only the bottleneck is synchronized during federated aggregation, the integration remains lightweight in terms of communication cost. 

This design can be applied to a variety of spatio-temporal architectures, including convolutional backbones such as ResNet~\cite{he2016deep} or MobileNet~\cite{howard2017mobilenets}, as long as they allow integration at the feature level.

\section{Evaluation}
\label{sec:eval}
\vspace{-2ex}

We evaluate the effectiveness of \framework{} in heterogeneous federated learning scenarios using the CIFAR-10 dataset. Our experimental setup consists of 50 clients, with 5 randomly selected participants per round. We simulate a non-IID setting by partitioning the dataset such that each client has, on average, 238.09 samples. Each round is trained with a learning rate of 0.0001 and a batch size of 64, for a total of 50 rounds. For benchmark, we compare \framework{} against three widely used FL baselines, FedAvg~\cite{mcmahan2017communication}, LG-FedAvg~\cite{liang2020think}, and FedProto~\cite{tan2022fedproto}. All comparison methods are evaluated under the same non-IID setting and with the same client models used in our framework. Section 4.1 evaluates how well \framework{} supports model heterogeneity using multiple backbone types, while Section 4.2 analyzes the effect of different design choices in the additive U-Net module using an ablation study.

\vspace{-2ex}
\subsection{Heterogeneous Model Support}

\begin{figure}[t]
    \centering
    \subfigure[Intra-family heterogeneity (VGG11, VGG16, and VGG19)]{
    \includegraphics[width=0.8\columnwidth]{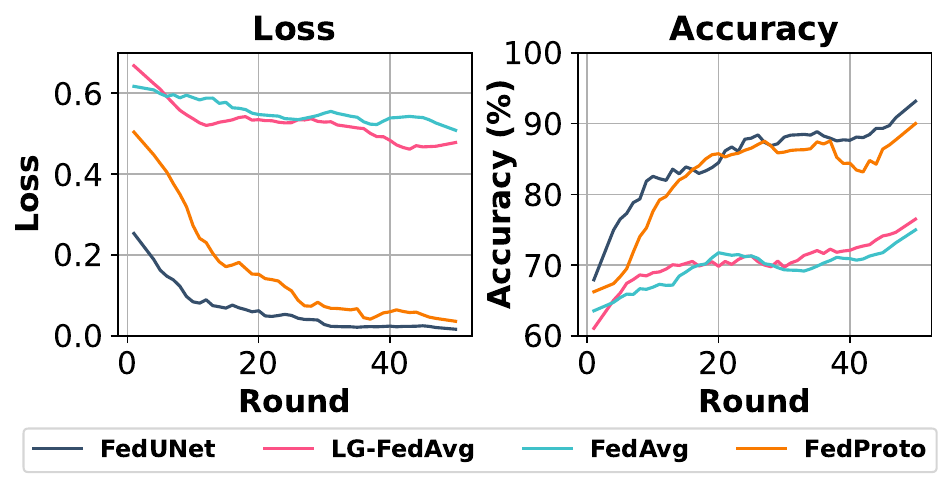}
    \label{fig:eval_intra}
    }
    \vspace{-2ex}
    \subfigure[Inter-family heterogeneity (VGG16 vs. ResNet18)]{
    \includegraphics[width=0.8\columnwidth]{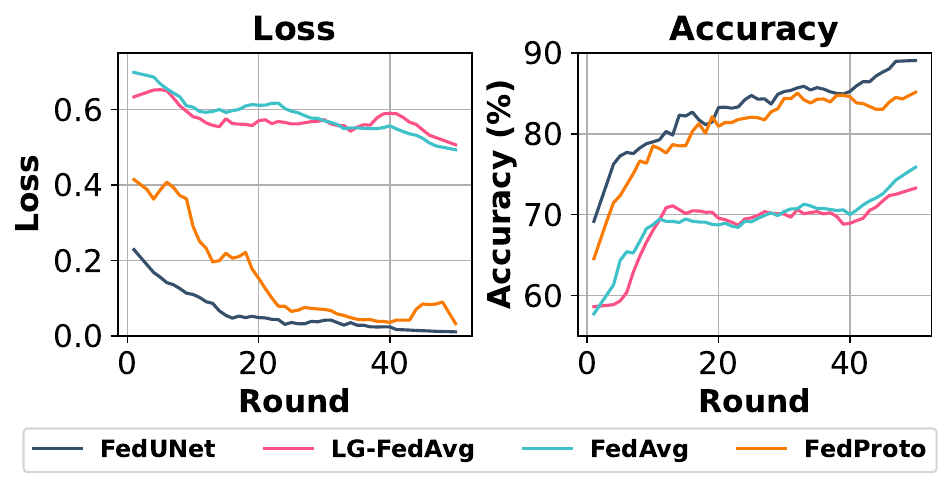}
    \label{fig:eval_inter}
    }
    \caption{Evaluation of heterogeneous model support.}
    \label{fig:eval_heterogeneity}
    \vspace{-2ex}
\end{figure}

To assess the ability of \framework{} to support structurally diverse client models, we conduct two types of experiments. First, we examine intra-family heterogeneity by assigning different variants of the VGG architecture (i.e., VGG11, VGG16, and VGG19) across clients. Second, we test inter-family heterogeneity by mixing VGG16 and ResNet18 across the federation.

For both settings, each client uses its own backbone model but shares a common additive U-Net module. During federated training, only the bottleneck parameters of the U-Net are synchronized among clients. We compare the performance of \framework{} with three baseline methods: FedAvg, LG-FedAvg, and FedProto, all of which are adapted to support heterogeneous models where applicable.

Our results show that \framework{} achieves competitive or superior accuracy compared to these baselines across heterogeneous configurations. Specifically, in the VGG11/16/19 setting, \framework{} maintains accuracy within \textbf{93.11}\% (a lightweight version of \framework{}: \textbf{92.68}\%) of the best-performing homogeneous baseline, while outperforming FedAvg, LG-FedAvg, and FedProto by 74.86\%, 76.16\%, and 90.03\%, respectively. In the more challenging VGG16/ResNet18 setting, \framework{} shows robust generalization and convergence, achieving \textbf{90.57}\% (a lightweight version of \framework{}: \textbf{89.32}\%) accuracy with significantly reduced communication cost compared to FedAvg and prototype-based aggregation.

These findings confirm that the shared bottleneck of the additive U-Net effectively harmonizes diverse feature representations, enabling seamless collaboration among heterogeneous clients without requiring model alignment or shared datasets.

\vspace{-2ex}
\subsection{Ablation Study on Additive U-Net Design}

\begin{figure}[t]
    \centering
    \includegraphics[width=0.8\columnwidth]{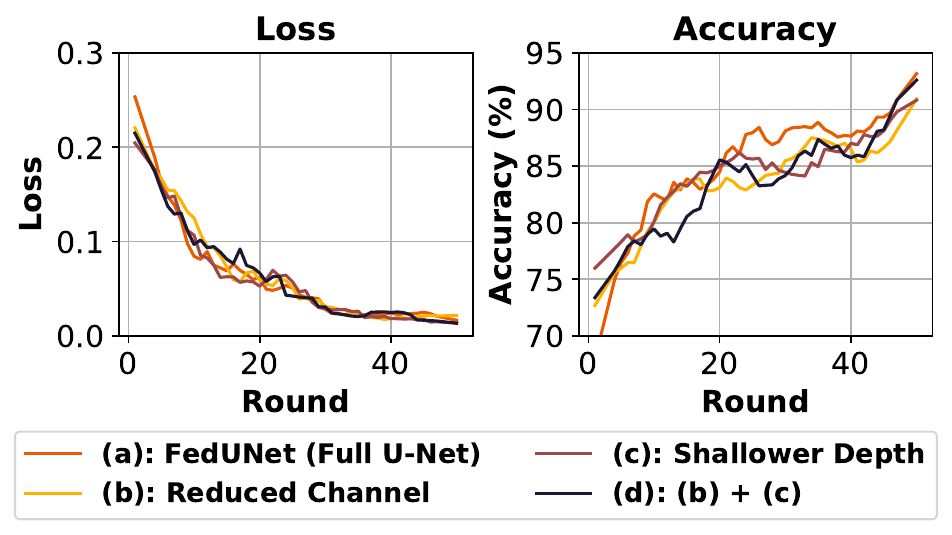}
    \vspace{-2ex}
    \caption{Ablation study on the different variants of the additive U-Net module using VGG16. Variants include (a) baseline additive U-Net, (b) reduced channels, (c) shallow depth, and (d) combined configuration with both reduced channels and shallow depth.}
    \label{fig:eval_ablation}
    \vspace{-3ex}
\end{figure}

To investigate how the design of the additive U-Net module influences performance, we conduct an ablation study using the VGG16 architecture for all clients. The study evaluates four different variants of the U-Net module.

\noindent{$\bullet$} (a) Full U-Net: The original encoder-decoder structure with 4 downsampling/upsampling layers and 64 channels per convolutional block.

\noindent{$\bullet$} (b) Reduced Channels: A lightweight version of (a) where all channels are reduced from 64 to 32 to minimize computation.

\noindent{$\bullet$} (c) Shallow Depth: A version of (a) where the depth of the encoder-decoder is reduced from 4 to 2, decreasing the number of downsampling operations.

\noindent{$\bullet$} (d) Reduced + Shallow: A compact version that combines both reduced channels and shallower depth.

All variants use the same training configuration. As before, only the bottleneck parameters are synchronized during federated updates. This evaluation isolates the impact of U-Net size and structure on both accuracy and communication efficiency.

Results show that the full U-Net design (a) achieves the highest accuracy at \textbf{93.92}\%, showing the effectiveness of deep, high-capacity modules in capturing transferable features. The combined variant (d), which reduces both channel width and depth, achieves slightly lower accuracy at \textbf{92.68}\%, indicating that lightweight setups can still retain much of the performance. The reduced channel (b) and shallow depth (c) variants yield comparable results, with accuracies of \textbf{91.96}\% and \textbf{91.79}\%, respectively. This indicates that both model width and depth contribute to representation quality.

These findings suggest a trade-off between model complexity and performance. While deeper and wider U-Nets are more effective, even compact variants can provide strong performance improvements over baselines by leveraging the shared bottleneck to learn client-invariant representations. Notably, these lightweight variants reduce communication cost from \textbf{56.66}MB to \textbf{0.89}MB with minimal accuracy degradation, making them practical for resource-constrained environments.
\vspace{-2ex}
\section{Discussions}
\label{sec:discussion}
\vspace{-2ex}

\subsection{Rationale behind employing U-Net}
While various encoder-decoder architectures could be considered for the additive module, we adopt U-Net due to its simplicity and effectiveness. U-Net’s structure with its encoder, decoder, and skip connections enables the extraction of both coarse and fine-grained features, making it a suitable choice for our goal of learning generalizable representations across heterogeneous clients. Furthermore, U-Net operates in a purely deterministic setting, unlike VAEs, which introduce stochastic latent variables and require additional loss terms such as KL-divergence. These properties make U-Net not only easier to train but also more stable and interpretable when integrated with standard supervised objectives like classification.

\vspace{-2ex}
\subsection{Skip Connection in Backbone-U-Net Fusion}
In \framework{}, the additive U-Net is fused with the backbone’s final feature map via a residual-like skip connection (addition or concatenation). This design allows the backbone to retain locally optimized features while the U-Net learns only complementary, client-invariant information. Like traditional residual learning, this reduces the learning burden on the U-Net, leading to more stable and efficient training. The skip connection enables cooperative learning between modules without disrupting gradients, supporting flexible integration across heterogeneous architectures.
\vspace{-2ex}
\section{Conclusion}
\label{sec:conc}
\vspace{-2ex}

We proposed \framework{}, a communication-efficient federated learning framework that supports heterogeneous model architectures via an additive U-Net module. By sharing only the bottleneck representation, our method enables client-invariant knowledge transfer without structural alignment. Experiments with diverse backbones demonstrate competitive performance and reduced communication cost. These results highlight \framework{} as a practical and scalable solution for federated learning in real-world heterogeneous settings.

\section*{Acknowledgements}
\vspace{-2ex}
The present research was supported by the research fund of Dankook University in 2024.




\bibliographystyle{elsarticle-num}
\bibliography{reference}


\end{document}